\documentclass[letterpaper, 10pt, conference]{ieeeconf}
\IEEEoverridecommandlockouts
\usepackage{cite}
\usepackage{amsmath,amssymb,amsfonts}
\usepackage{algpseudocode}
\usepackage{algorithm}		
\usepackage{graphicx}
\usepackage{textcomp}
\usepackage{xcolor}
\usepackage{fancyhdr}
\DeclareMathOperator*{\argmax}{argmax}

\begin{document}
\bstctlcite{IEEEexample:BSTcontrol}

\title{Filter Early, Match Late: Improving Network-Based Visual Place Recognition}

\author{Stephen Hausler, Adam Jacobson and Michael Milford
\thanks{
The work of S. Hausler was supported by a Research Training Program Stipend. The work of M. Milford was supported by the Australian Research Council Future Fellowship under Grant FT140101229.
The authors are with the ARC Centre of Excellence for Robotic Vision, Queensland University of Technology, Brisbane, QLD 4000, Australia (e-mail: stephen.hausler@hdr.qut.edu.au).}
}

\maketitle
\thispagestyle{fancy}

\pagestyle{empty}

\begin{abstract}
CNNs have excelled at performing place recognition over time, particularly when the neural network is optimized for localization in the current environmental conditions. In this paper we investigate the concept of feature map filtering, where, rather than using all the activations within a convolutional tensor, only the most useful activations are used. Since specific feature maps encode different visual features, the objective is to remove feature maps that are detract from the ability to recognize a location across appearance changes. Our key innovation is to filter the feature maps in an early convolutional layer, but then continue to run the network and extract a feature vector using a later layer in the same network. By filtering early visual features and extracting a feature vector from a higher, more viewpoint invariant later layer, we demonstrate improved condition and viewpoint invariance. Our approach requires image pairs for training from the deployment environment, but we show that state-of-the-art performance can regularly be achieved with as little as a single training image pair. An exhaustive experimental analysis is performed to determine the full scope of causality between early layer filtering and late layer extraction. For validity, we use three datasets: Oxford RobotCar, Nordland, and Gardens Point, achieving overall superior performance to NetVLAD. The work provides a number of new avenues for exploring CNN optimizations, without full re-training. 

\end{abstract}

\section{Introduction}

Convolutional neural networks have demonstrated impressive performance on computer vision tasks \cite{AlexNet,He}, including visual place recognition \cite{SN2015,AR2018}. Recently, researchers have investigated optimising and improving pre-trained CNNs, by either extracting salient features \cite{Lost2018,Zetao2018}, or by `pruning' the network \cite{FeatPruneFirst2017,FeatPrune2018}. Network pruning is typically used to increase the computation speed of forward-pass computation; however, our previous work has provided a proof of concept that a type of pruning, dubbed ``feature map filtering", can \textit{also} improve the place recognition performance of a pre-trained CNN \cite{Hausler}. 

\begin{figure}[t]
\centering
\includegraphics[width=0.95\linewidth,trim=1cm 11.5cm 1cm 1cm,clip]{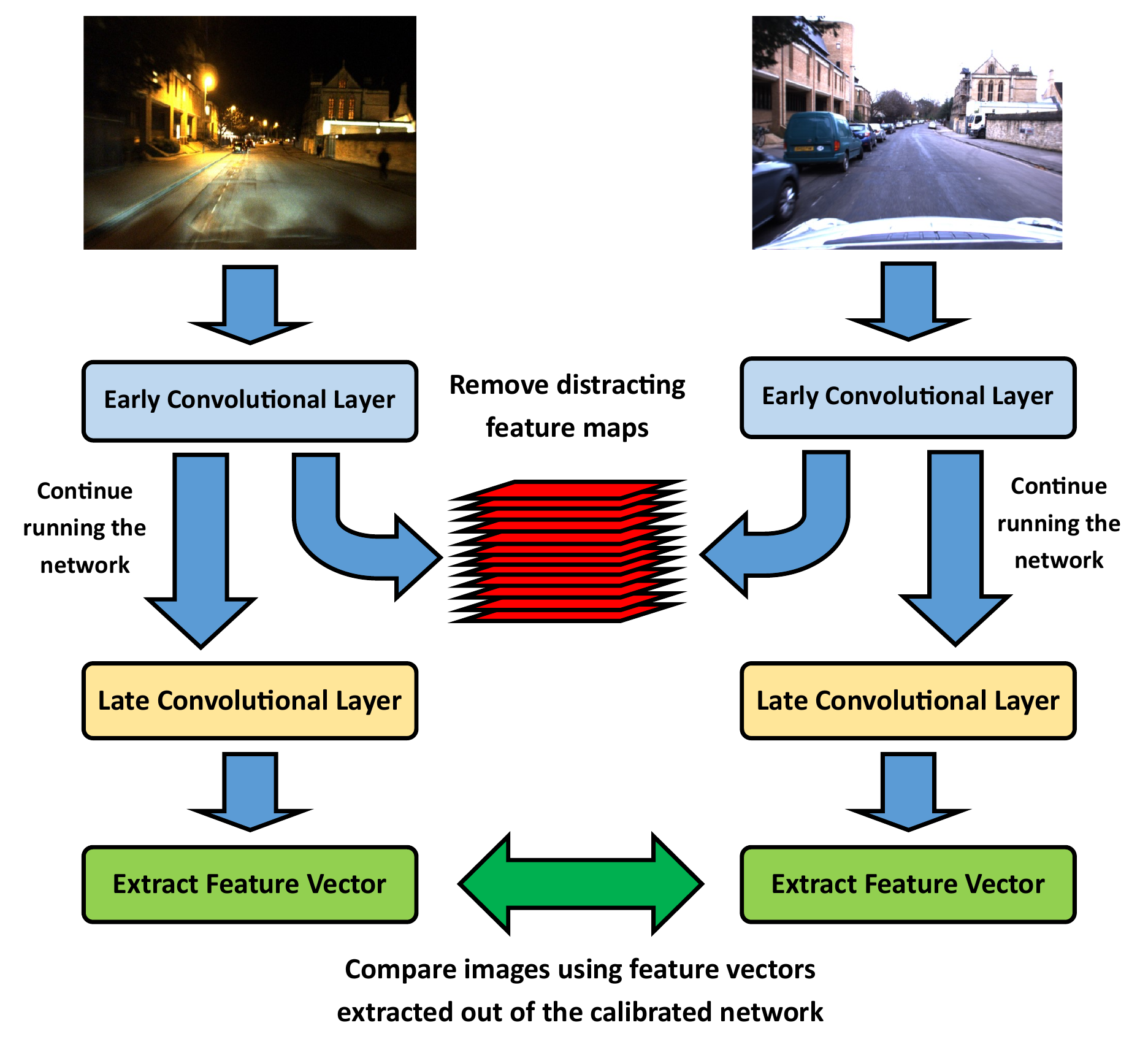}	
\caption{Our method removes early layer feature maps that have learnt visual features that detract from place recognition performance to improve matching performance using downstream layers.}
\label{Placeholder}
\end{figure}

In feature map filtering, specific feature maps are removed, based on their suitability to identify the correct matching location across a changing environmental appearance. In this work, we propose that an early convolutional layer can be filtered to improve the matching utility of feature vectors extracted from a network's \textit{later} layers. By performing this early layer filtering, simple visual features (e.g. textures and contours) that detract from a network's utility for place recognition across changing environmental conditions are removed. Crafting a feature vector out of a later layer is beneficial, as research \cite{GS2018,SN2015} has shown that later CNN layers are more invariant to viewpoint changes. We verify the ability to handle viewpoint variations by using the Gardens Point Walking dataset, and handle condition variations using the Oxford RobotCar dataset (matching from night to day) and the Nordland dataset (matching from summer to winter). 

We summarize the contributions of this work:
\begin{itemize}
\item{We propose a novel method of performing feature map filtering (or pruning) on early convolutional layers, while extracting features for place recognition out of later convolutional layers.}
\item{To determine the selection of feature maps to remove, we have developed a Triplet Loss calibration procedure which uses training image pairs to remove feature maps that show consistent detriment in the ability to localize in the current environment. We demonstrate experimentally that state-of-the-art performance can be achieved with as little as a single training image pair.}
\item{We provide a thorough experimental evaluation of the effects of filtering CNN feature maps for a pre-trained neural network, exhaustively testing all combinations in the layers Conv2 to Conv5. We also include a set of experiments filtering Conv2 and using the first fully connected layer as a feature vector.}
\end{itemize}
Our results also reveal the inner workings of neural networks - a neural network can have a portion of it's feature maps completely removed and yet a holistic feature vector can be extracted out of a higher convolutional layer. We also provide a visualization of the activations within a higher layer of the filtered network.

The paper proceeds as follows. In Section II, we review feature map pruning literature and discuss the application of neural networks in visual place recognition. Section III presents our methodology, describing the calibration procedure. Section IV details the setup of our three experimental datasets and Section V discusses the performance of filtering different convolutional layers on these datasets. Section VI summarizes this work and provides suggestions for future work.

\section{Related Work}

The recent successes of deep learning in image classification \cite{IM2015} and object recognition \cite{COCO} have encouraged the application of neural networks in place recognition. In early work, the pre-trained AlexNet \cite{AlexNet} network is used to produce a feature vector out of the Conv3 layer \cite{SN2015,Sunderhauf2015}. Rather than simply using a pre-trained network, NetVLAD learns visual place recognition end-to-end. In NetVLAD, triplet loss is used to find the optimal VLAD encoding to match scenes across both viewpoint and condition variations \cite{AR2018}. LoST uses the semantic CNN RefineNet \cite{Lin2016} to select salient keypoints within the width by height dimensions of a convolutional tensor \cite{Lost2018}. In a related work, these keypoints have been found by observing the activations out of a late convolutional layer \cite{LookOnce2017}. The aforementioned examples involve improving a pre-trained neural network for place recognition, either by re-training, or selecting the most useful components out of the network activations. 

Several works perform network simplification. Initially, CNNs were compressed by selectively removing specific convolutional weights. For example, Han et al. prunes convolutional weights that are below a certain threshold, then applies quantization and Huffan coding to reduce the network size by up to 49 times \cite{Han2015}. An alternate strategy is to remove all weights in an entire feature map, which is termed filter pruning. Li et al. uses the absolute magnitude of weights within each feature map as the pruning metric, with low magnitude maps being removed \cite{Li}. An alternate approach is to perform Linear Discriminant Analysis, to preserve the class structure in the final layer while reducing the number of feature maps in earlier layers \cite{FeatPrune2018}. A recent work, which is currently under review, uses structured sparsity regularization and Stochastic  Gradient  Descent to prune unnecessary feature maps \cite{Lina}. A `soft' filter has been used to enhance visual place recognition, where the activations across a stack of feature maps are scaled with a learnt scaling value and summed together to create a learnt feature vector. There are two works in this space: the first performs feature weighting on the last convolutional layer \cite{Noh}, while the second re-weights a concatenation of features out of multiple convolutional layers \cite{Mao2018}.

In the aforementioned feature map pruning literature, after filtering feature maps, the smaller network is briefly re-trained (to remove sparsity). In this work, we show that a network can be left sparse ``as-is" and continues to produce coherent activations in higher network layers, even up to the fully-connected layer.

\section{Proposed Approach}

A pre-trained neural network, which was trained on a diverse set of images, will learn internal representations of a wide range of different visual features. However, in visual place recognition, perceptual aliasing is a common problem. Perceptual aliasing is where certain visual features make a scene visually similar to a previously observed scene in a different location. If a pre-trained network is selectively filtered on the expected environment, then visual features that contribute to perceptual aliasing can be removed, leaving the feature maps that encode visual features that can suitably match between two appearances of the same location. We use a short calibration method to prepare our feature map filter, as described in the following sub-sections.

\subsection{Early Feature Filtering, Late Feature Extraction}

In our previous work on feature map filtering \cite{Hausler}, we filtered the feature map stack while extracting the feature vector from the same layer. While we demonstrated improved place recognition performance, this approach was not capable of optimizing for the extraction of visual features in higher convolutional layers. We hypothesize that filtering an early convolutional layer will remove distracting visual features, while crafting a feature vector out of a later layer has been shown \cite{GS2018} to have improved viewpoint robustness. 

Our improved approach filters the feature map space within a CNN, except the network is allowed to continue after filtering. Optimizing the filter is now dependent on the triplet loss on the features extracted out of a higher convolutional layer. This also adds the concept of feedback to a neural network, by modulating early visual features with respect to higher level features. 

\subsection{Deep Learnt Feature Extraction and Triplet Loss}

A feature vector is extracted out of a width by height by channel ($W \times H \times C$) convolutional tensor. To improve viewpoint robustness and increase the processing speed, dimensionality reduction is performed using spatial pooling \cite{CZ2017}. As per our previous work \cite{Hausler}, we again use pyramid spatial pooling and convert each $W \times H$ tensor into a vector of dimension 5, containing the maximum activation across the entire feature map plus the maximum activation in each quadrant of the feature map. For all our experiments we use the pre-trained network HybridNet \cite{CZ2017}. HybridNet was trained with place recognition in mind, resulting in a well-performing pre-trained network with a fast forward-pass processing speed.

\begin{figure}[t]
\centering
\includegraphics[width=0.9\linewidth,trim=2cm 2cm 1.5cm 2cm,clip]{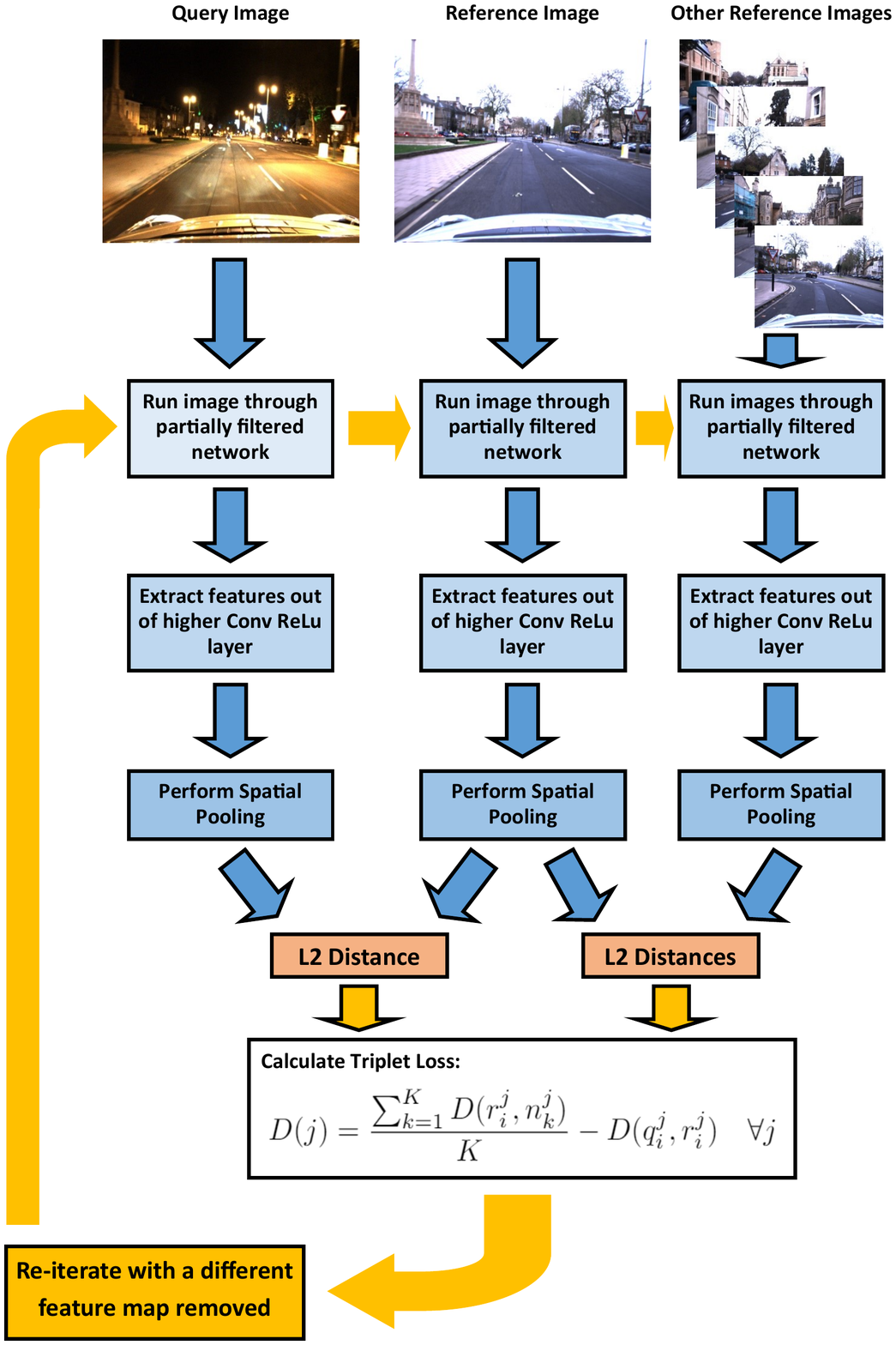}	
\caption{Flow chart showing the Triplet Loss implementation used to rank the importance of early layer feature maps.}
\label{FeatFiltMethod}
\end{figure}

Triplet Loss \cite{AR2018} involves comparing the feature distance between a positive pair of images relative to one or more negative pairs of images. In this case, the positive pair are two images taken from the same physical location, but at different times, while the negative pairs are images taken at a similar time but at varying locations. We use one `hard' negative pair, in this case, the second reference image is an image that is a fixed number of frames ahead of the current frame. This distance is slightly larger than the ground truth tolerance. We then have four `soft' negatives, which are random images elsewhere in the reference dataset. Including a fixed, hard negative reduces variance in the filter calibration. As per literature best practise \cite{AR2018,Loquercio2017}, we use the L2-Distance as our optimization metric. Figure \ref{FeatFiltMethod} shows an overview of our proposed approach.

\subsection{Filtering Method}

We use a type of Greedy algorithm \cite{Fegaras1998} to determine which subset of the feature map stack suits the current environmental conditions. Our variant of the Greedy algorithm finds the worst performing (with respect to the triplet loss) feature map at each iteration of the algorithm. Normally the Greedy algorithm will terminate when the local minimum triplet loss is reached; however, to guarantee that the global minimum is found, we continue iterating until half the original feature map stack has been removed. We store the filter selection at each iteration and search for the global minimum across all iterations. 

Additionally, we also implement a batch filtering extension to the aforementioned algorithm. In batch filtering, in each iteration, the four worst feature maps are discovered (based on the triplet loss) and removed before the next iteration. We can safely add this approximation because of the global minimum search. If removing four maps at once prevents the loss function from being convex, the best match can still be found due to the global search. Adding batch filtering improves the computation speed of calibration by a factor of four. The decision to terminate the search at half the maps removed is a heuristically determined trade-off between calibration processing time and localization benefits. 

Determining the worst performing feature map is based on the triplet loss score out of a higher network layer. Specifically, each feature map is individually removed and the network is continued to run further into the forward pass. The triplet loss is then calculated based on the feature vectors extracted out of a higher network layer. As mentioned previously, we apply a maximum spatial pooling operation on the raw Conv ReLu activations. The purpose of this is to reduce the dimensionality of the feature vector, and to ensure the filtering process focuses on strong activations. For each pair of images in the triplet set, the L2 distance between that pair is calculated as per the equation below.

\begin{equation}
	D(q_i^j,r_i^j) = \sqrt{ \sum^M_{k=1} (q_i^j(k) - r_i^j(k))^2}
\end{equation}
\noindent where $M$ is the dimension of the filtered query feature vector $q_i^j$.

The equation above is repeated for the five negative pairs. The difference scores for the five pairs are then averaged together. The triplet loss is the difference between the positive pair and the averaged negative pair, across a different feature map $j$ being removed.

\begin{equation}
    D(j) = \frac{\sum_{k=1}^{K} D(r_i^j,n_k^j)}{K} - D(q_i^j,r_i^j)\quad \forall j
\end{equation}
\noindent where $r_i^j$ represents the current location filtered reference feature vector and $n_i^j$ represents the averaged negative. $j$ denotes the index of the currently filtered feature map. In our experiments, K is set to 5 since the set of $n_k$ consists of one fixed reference image and four, randomly selected, reference images.

We then find the maximum distance:
\begin{equation}
    maxval = \underset{1 \leq j \leq N}{\max} D(j)
\end{equation}
\begin{equation}
    worstFmap = \underset{1 \leq j \leq N}{\argmax}\:\: D(j)  \label{worstmap}
\end{equation}
\noindent where N is the number of remaining feature maps.

The index of the maximum distance represents the feature map to be removed to achieve the greatest L2 difference between the images from the same location and the average negative distance. To implement the batch filtering, the previous worst map from $D$ is removed and equation \ref{worstmap} is repeated until the four worst feature map ids are collected.

At the end of each iteration, the weights and biases in an earlier convolutional layer are set to zero, for each weight inside the feature maps selected to be filtered. We then return the new, partially zeroed CNN to the next iteration of the algorithm. These iterations continue until half the features maps are removed. The global maximum triplet loss score is then found, and the selection of filtered feature maps at the maximum loss score are the final set of filtered maps for that calibration image. 

Finally, for improved robustness and to prevent outliers, we use multiple calibration images. The choice of filtered feature maps is stored for all images and after the calibration procedure is finished, the number of times a particular feature map is removed is summed across all the calibration image sets. The final filtered maps are feature maps that were chosen to be removed in at least 66\% of the calibration sets. This threshold was heuristically determined, using place recognition experimentation on a range of different thresholds. With this threshold, on average, approximately 25\% of earlier layer feature maps are removed after filtering. We chose this metric based on the objective to find feature maps that are consistently poor at navigation, rather than feature maps that are only inefficient for a single image. This approach reduces the risk of overfitting the filter to the calibration data.

\subsection{Place Recognition Filter Verification Algorithm}

To evaluate the performance of the calibrated feature map filter, we use a single-frame place recognition algorithm. To apply the filter, every convolutional weight in a filtered feature map is set to zero. This new network is then run in the forward direction to produce convolutional activations in higher network layers. We again apply spatial pooling to the convolutional activations, producing a feature vector of length five times the number of feature maps.

The feature vectors from the reference and query traverses are compared using the cosine distance metric. While the euclidean distance was the training distance metric, our experiments revealed that it is advantageous to train the filter using the euclidean distance metric but perform place recognition using the cosine similarity metric. In early experiments, we also checked the performance by training with the cosine distance metric instead and found a reduction in the resulting place recognition performance. 

The resultant difference vector is then normalized to the range 0.001 to 0.999, where 0.001 denotes the worst match and 0.999 the best match. We then apply the logarithm operator to every element of the difference vector. Taking the logarithm amplifies the difference between the best match and other, perceptually aliased matches \cite{Hausler2019}. The place recognition quality score is calculated using the method originally proposed in SeqSLAM \cite{MM2012}, where the quality score is the ratio between the score at the best matching template and the next largest score outside a window around the best matching template. A set of precision and recall values are calculated by varying a quality score threshold value. For compact viewing, we display the localization performance using the maximum F1 score metric, where the F1 score is the harmonic mean of precision and recall. 

\section{Experimental Method}

We demonstrate our approach on three benchmark datasets, which have been extensively tested in recent literature \cite{GS2018,SRAL,NT2018}. The datasets are Oxford RobotCar, Nordland, and Gardens Point Walking. Each dataset is briefly described in the sections below.

\textbf{Oxford RobotCar} - RobotCar was recorded over a year across different times of day, seasons and routes \cite{RobotCar}. For our training set, we use 50 image sets (a positive image, an anchor and five negative images) extracted at an approximate frame rate of one frame every two seconds. Using a low frame rate ensures that the individual images show some diversity between them. Therefore, the calibration set has a duration of approximately 100 seconds, which is a realistic and practical calibration duration for a real-world application. We also experiment with a smaller number of calibration image sets, to observe the effects of using fewer calibration images.

For our test set, we use 1600 frames extracted out of the dataset, which corresponds to approximately two kilometers through Oxford. There are no training images present in the test set. The reference dataset was recorded on an overcast day (2014-12-09-13-21-02), while the query dataset is at nighttime on the following day (2014-12-10-18-10-50). We use a ground truth tolerance of 30 meters, consistent with recent publications \cite{GS2018,Hausler2019}.

\textbf{Nordland} – The Nordland dataset \cite{NordlandDatasetRef} is recorded from a train travelling for 728 km through Norway across four different seasons. The training set again consists of 50 images, with a recording frame rate of 0.2 frames per second. The resultant calibration duration is 250 seconds; a longer real-world duration was heuristically chosen to account for the significantly larger real-world distance of the Nordland dataset (compared to Oxford RobotCar or Gardens Point Walking). 

For the experimental dataset, we use the Winter route as the reference dataset and the Summer traverse as the recognition route, using a 2000 image subset of the original videos. In our previous work \cite{Hausler} we used the Summer images as the reference set and the Winter images as query; we flipped the order because we found that matching from Summer to Winter to be more challenging. For the ground truth we compare the query traverse frame number to the matching database frame number, with a ground-truth tolerance of 10 frames, since the two traverses are aligned frame-by-frame. Again the test set contains no images from the training set.

\textbf{Gardens Point Walking} - was recorded at the QUT university campus in Brisbane and consists of two traverses during the day and one at night, with a duration of 200 images per traverse \cite{SN2015}. One of the day traverses is viewed from the left-hand side of the walkways, while the second day and the night traverse were both recorded from the right-hand side. We train our filter on the comparison between the left-hand side at daytime to the right-hand side at nighttime, using just 5 calibration images. We then use 194 images as the evaluation set and a ground truth tolerance of 3 frames.

\section{Results}

In this section, a detailed analysis is performed on the performance of feature map filtering in visual place recognition. The results are shown using the maximum F1 score metric and we compare our early layer filter approach to three benchmarks. First, we compare against filtering the same layer as the feature vector is extracted from. To ensure a fair comparison, the same triplet loss method is used, including using five negative images. The second benchmark is the localization performance without any filtering at all. Finally, we also compare against pre-trained NetVLAD (trained on Pittsburgh 30k) \cite{AR2018}. NetVLAD normally outputs results as a Recall@N metric; we convert this to an equivalent F1 score by assuming a precision score of 100\% and using the Recall@1 value as the recall score. 

\begin{figure}[h]
\centering
\includegraphics[width=0.9\linewidth,trim=2cm 2cm 2cm 0cm,clip]{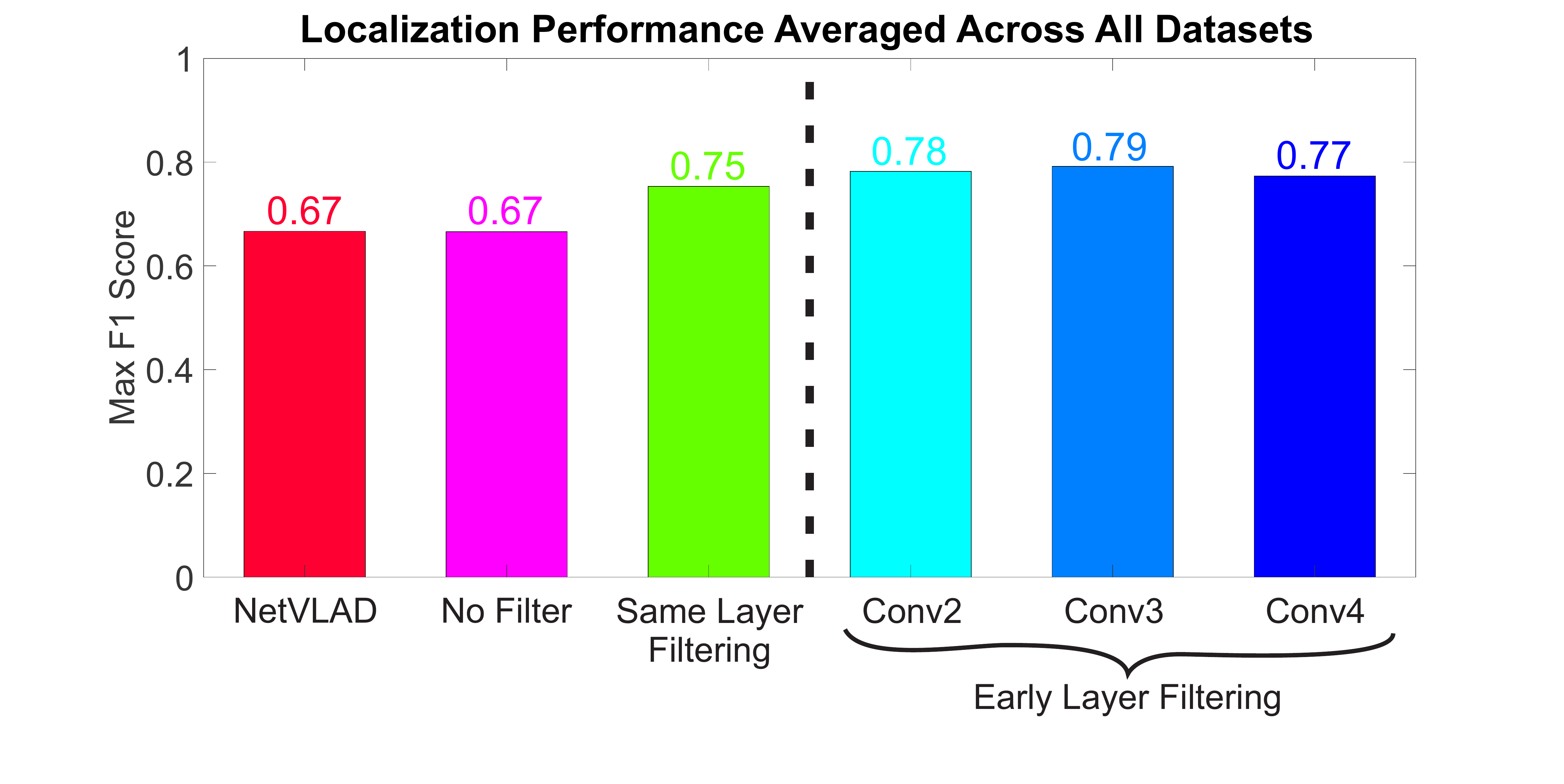}	
\caption{Overall performance with feature map filtering, averaged across the three datasets.}
\label{Overall}
\end{figure}

Figure \ref{Overall} provides a summary of the overall place recognition performance across all three datasets. Overall, removing feature maps in the same layer as the feature vector is extracted from has a higher maximum F1 score with respect to both NetVLAD, and the same network without any filtering. Filtering the feature maps in an earlier convolutional layer produces a further improvement to the average place recognition performance.

\subsection{Oxford RobotCar}

\begin{figure}[h]
\centering
\includegraphics[width=0.9\linewidth,trim=2cm 3cm 2cm 0cm,clip]{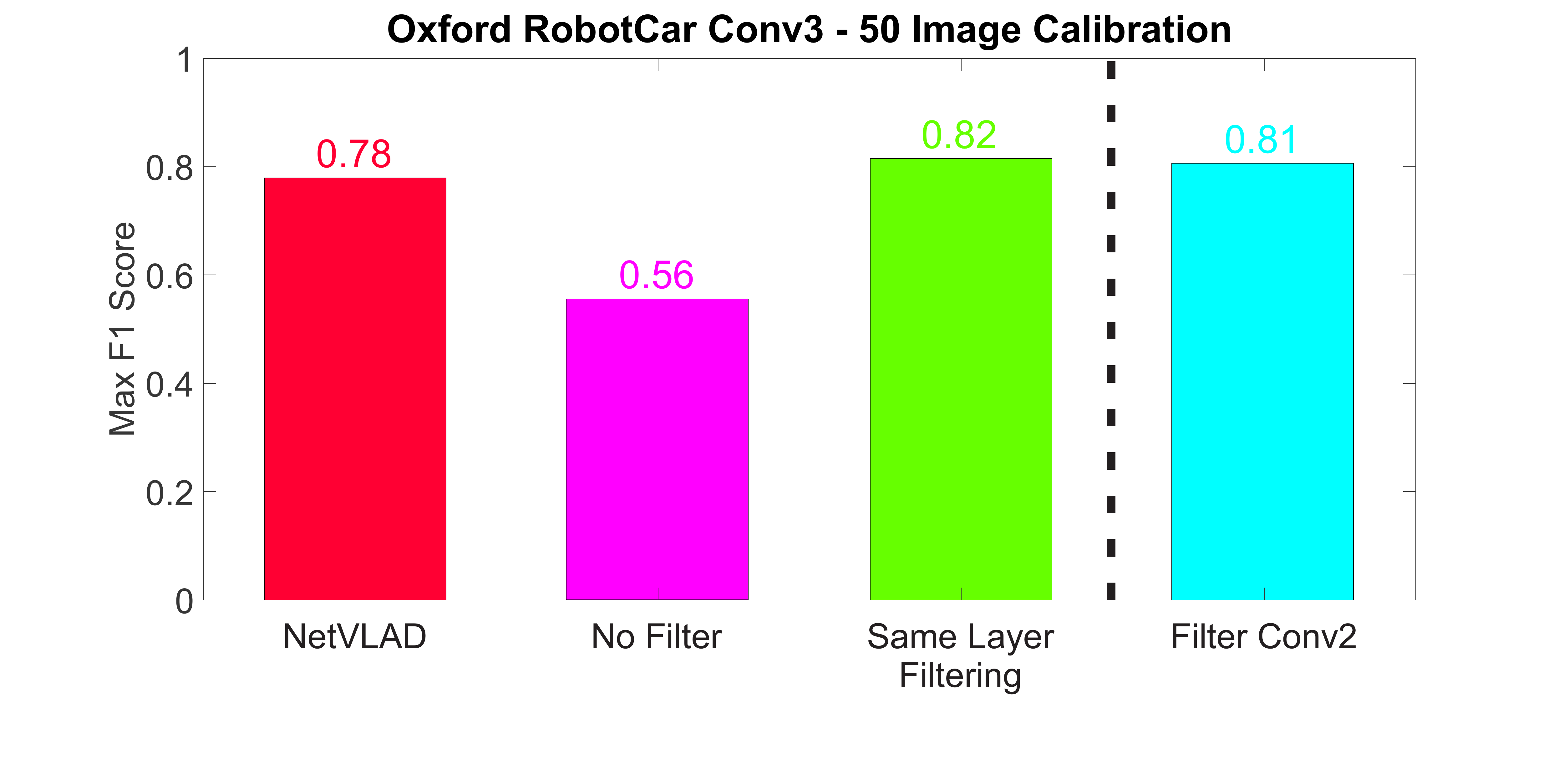}	
\caption{Maximum F1 score for Feature Map Filtering on the Oxford RobotCar dataset, extracting features from the Conv3 layer. We compare the localization performance without any filtering (\emph{No Filter}), with filtering the same layer from which the image descriptor is extracted (\emph{Same Layer Filtering}), and filtering the previous layer (\emph{Filter Conv2}). Finally we compare our results to NetVLAD.}
\label{F1BarOxfordConv3}
\end{figure}

\begin{figure}[h]
\centering
\includegraphics[width=0.9\linewidth,trim=2cm 1.5cm 2cm 0cm,clip]{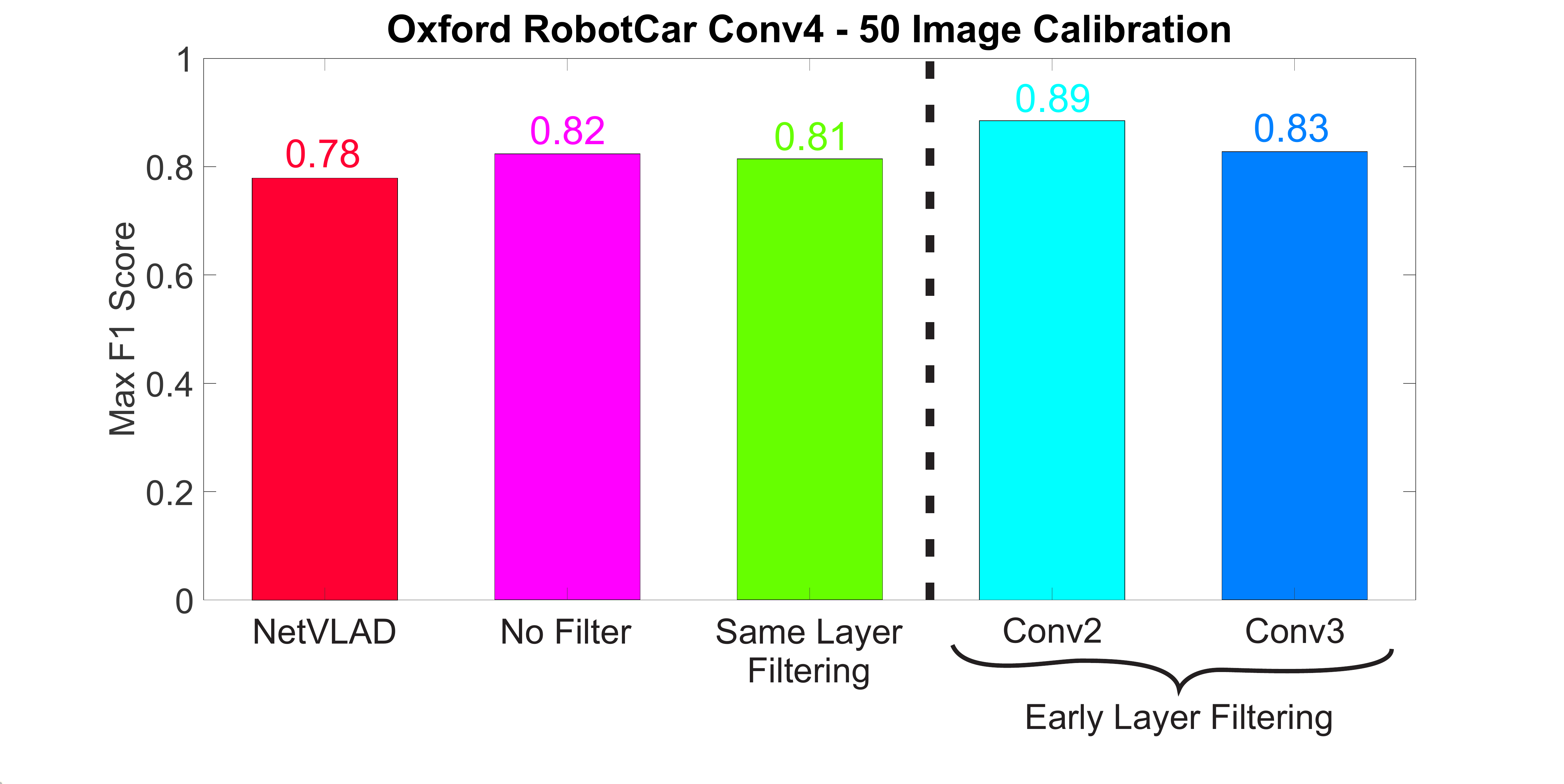}	
\caption{Maximum F1 score for Feature Map Filtering on the Oxford RobotCar dataset, extracting features from Conv4. Again we compare the localization performance without filtering (\emph{No Filter}), with filtering the same layer (\emph{Same Layer Filtering}), and filtering either of the two previous layers (\emph{Early Layer Filtering}). Finally we compare our results to NetVLAD.}
\label{F1BarOxfordConv4}
\end{figure}

\begin{figure}[h]
\centering
\includegraphics[width=0.9\linewidth,trim=2cm 1.5cm 2cm 0cm,clip]{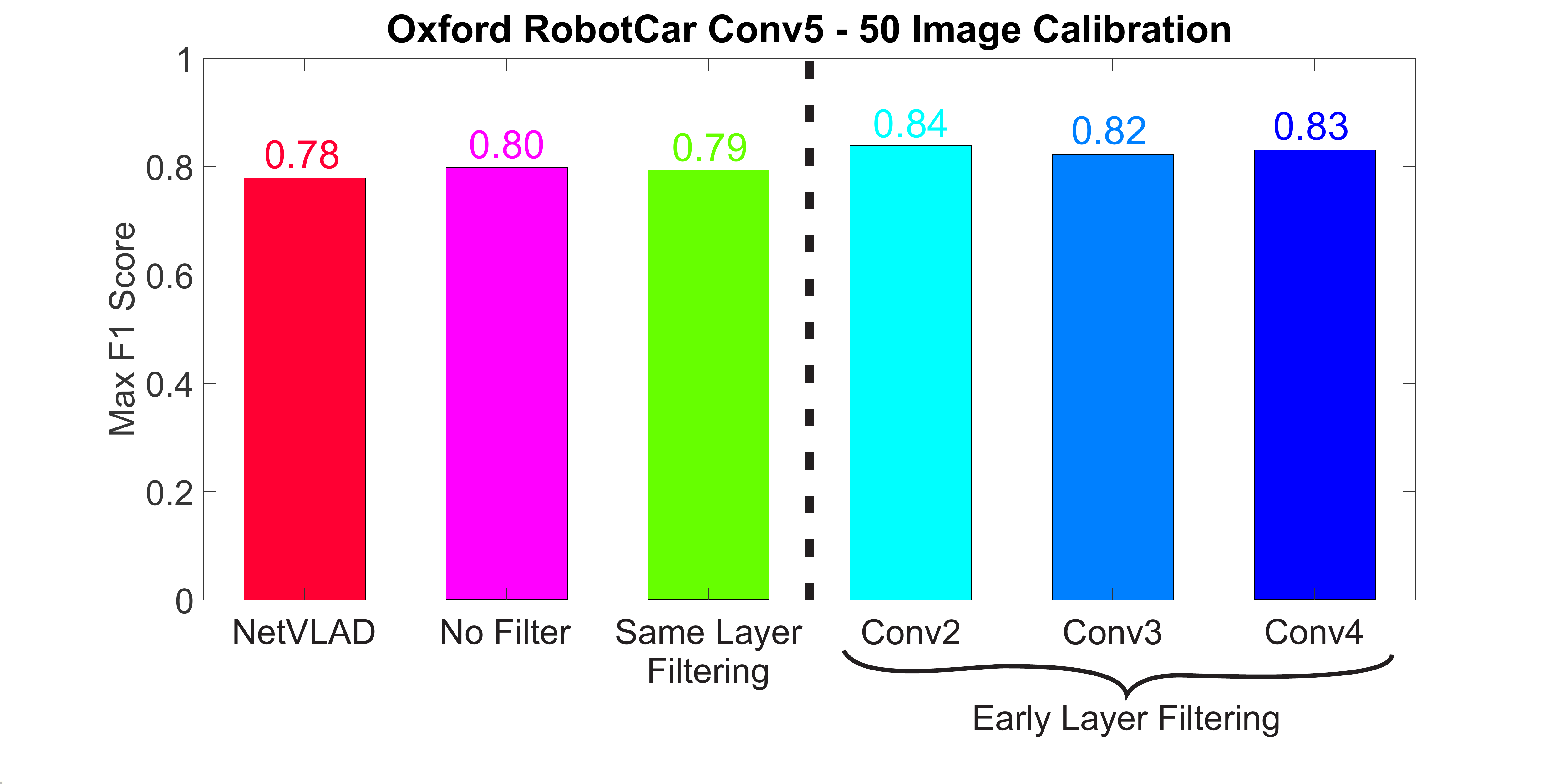}	
\caption{Maximum F1 score for Feature Map Filtering on the Oxford RobotCar dataset, extracting features from the Conv5 layer. We compare the localization performance without any filtering (\emph{No Filter}), with filtering the same layer from which the image descriptor is extracted (\emph{Same Layer Filtering}), and filtering either of the three previous layers (\emph{Early Layer Filtering}). Finally we compare our results to NetVLAD.}
\label{F1BarOxfordConv5}
\end{figure}

Early layer filtering generally improves localization on the Oxford RobotCar dataset (see Figures \ref{F1BarOxfordConv3} to \ref{F1BarOxfordConv5}). If Conv3 features are used for localization, then whether an earlier layer or the current layer is filtered is largely irrelevant. However, whichever method is used results in a significant improvement in localizing with these features. When Conv2 is filtered and Conv4 features are used, the localization experiment results in a maximum F1 score improvement of 0.8, compared to filtering on the same convolutional layer (Conv4). 

\begin{figure}[p]
\centering
\includegraphics[width=\linewidth,trim=2cm 0cm 2cm 0.3cm,clip]{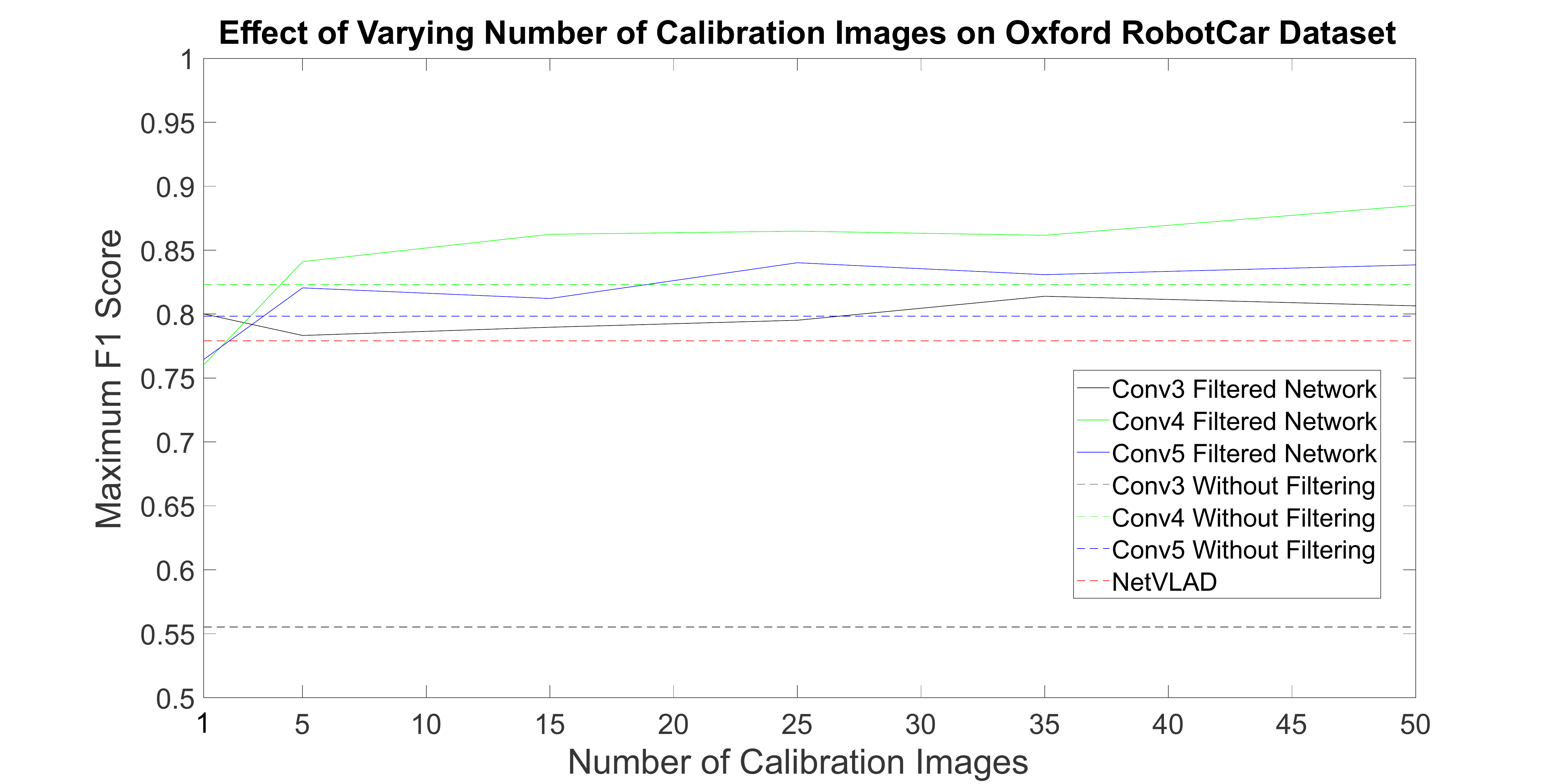}	
\caption{In this experiment, we filter feature maps from the Conv2 layer and extract a feature vector out of one of the later layers. Increasing the number of calibration images provides a small improvement to the localization ability of the network. Even if just 5 calibration images are available, the filtering approach still beats both the baseline without filtering, and NetVLAD. With Conv3 features and a single calibration image, the maximum F1 score is higher than NetVLAD.}
\label{Conv2_Dynamic_Calibration}
\end{figure}

In Figure \ref{Conv2_Dynamic_Calibration}, we varied the number of calibration images used when training the filter on the Conv2 layer. We used as little as 1 calibration image, up to 50 calibration images. We determined that the localization performance improves gradually, and even calibrating with five images in the query environment improves the place recognition performance above both NetVLAD and HybridNet without any filtering. This is particularly apparent with Conv3 features, which normally are not a suitable choice for a localization system. As a general rule, the more calibration images, this lower the risk of over-fitting the filter on the calibration data. These results indicate that even if only a single calibration image is available, our approach can provide an improvement to localization. This also indicates that there are visual features which are a detriment to place recognition across all variations in the remainder of the Oxford RobotCar dataset (from night to day), such that 1 image of the environment is sufficient to remove many of these poor visual features.

\clearpage

\begin{figure}[p]
\centering
\includegraphics[width=\linewidth,trim=2cm 3cm 2cm 0cm,clip]{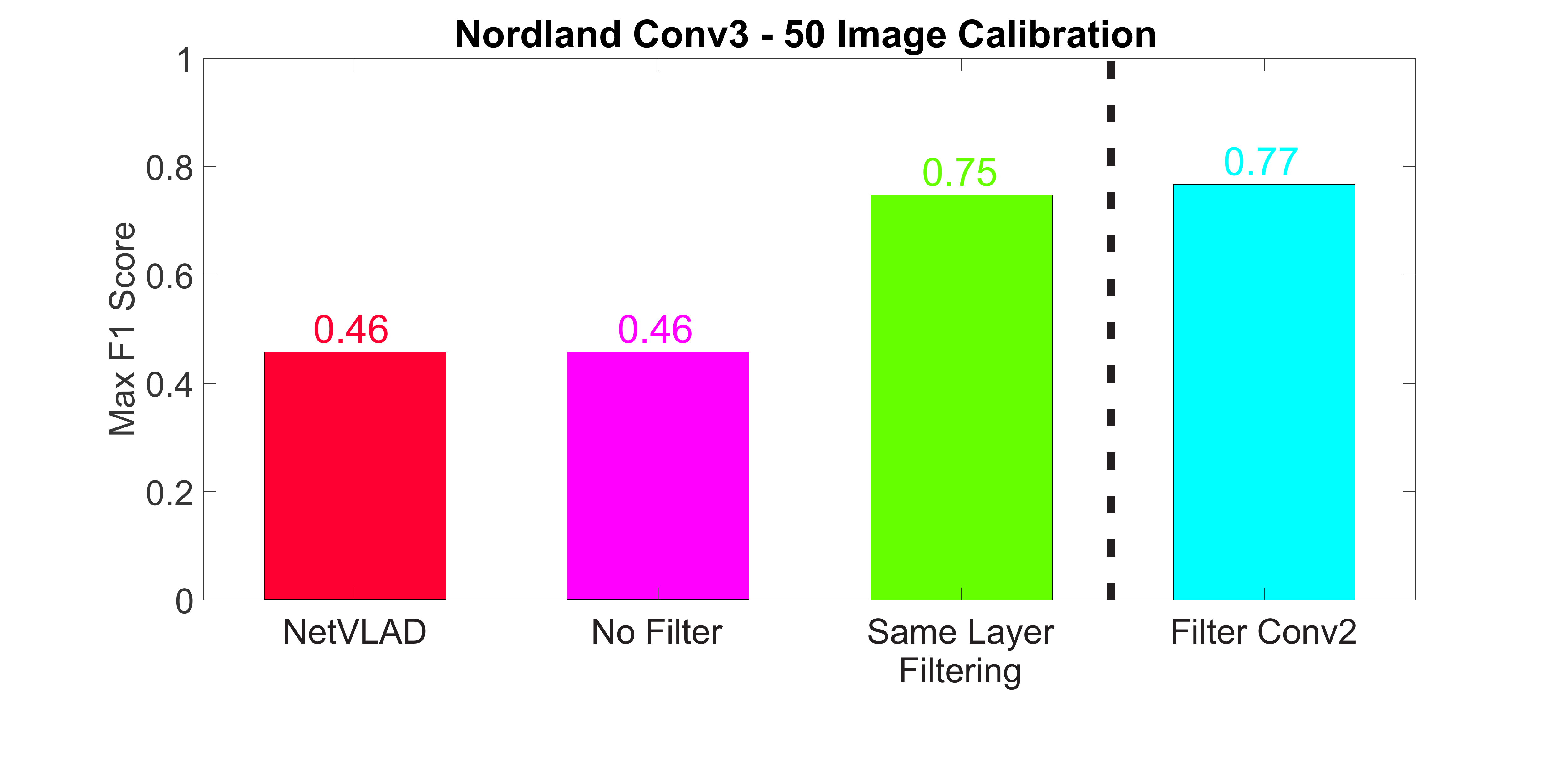}	
\caption{Maximum F1 score for Feature Map Filtering on the Nordland dataset, extracting features from Conv3. We compare the localization performance without any filtering (\emph{No Filter}), with filtering the same layer from which the image descriptor is extracted (\emph{Same Layer Filtering}), filtering the previous layer (\emph{Filter Conv2}), and to NetVLAD.}
\label{F1BarNordConv3_50}
\end{figure}

\begin{figure}[p]
\centering
\includegraphics[width=\linewidth,trim=2cm 1.5cm 2cm 0cm,clip]{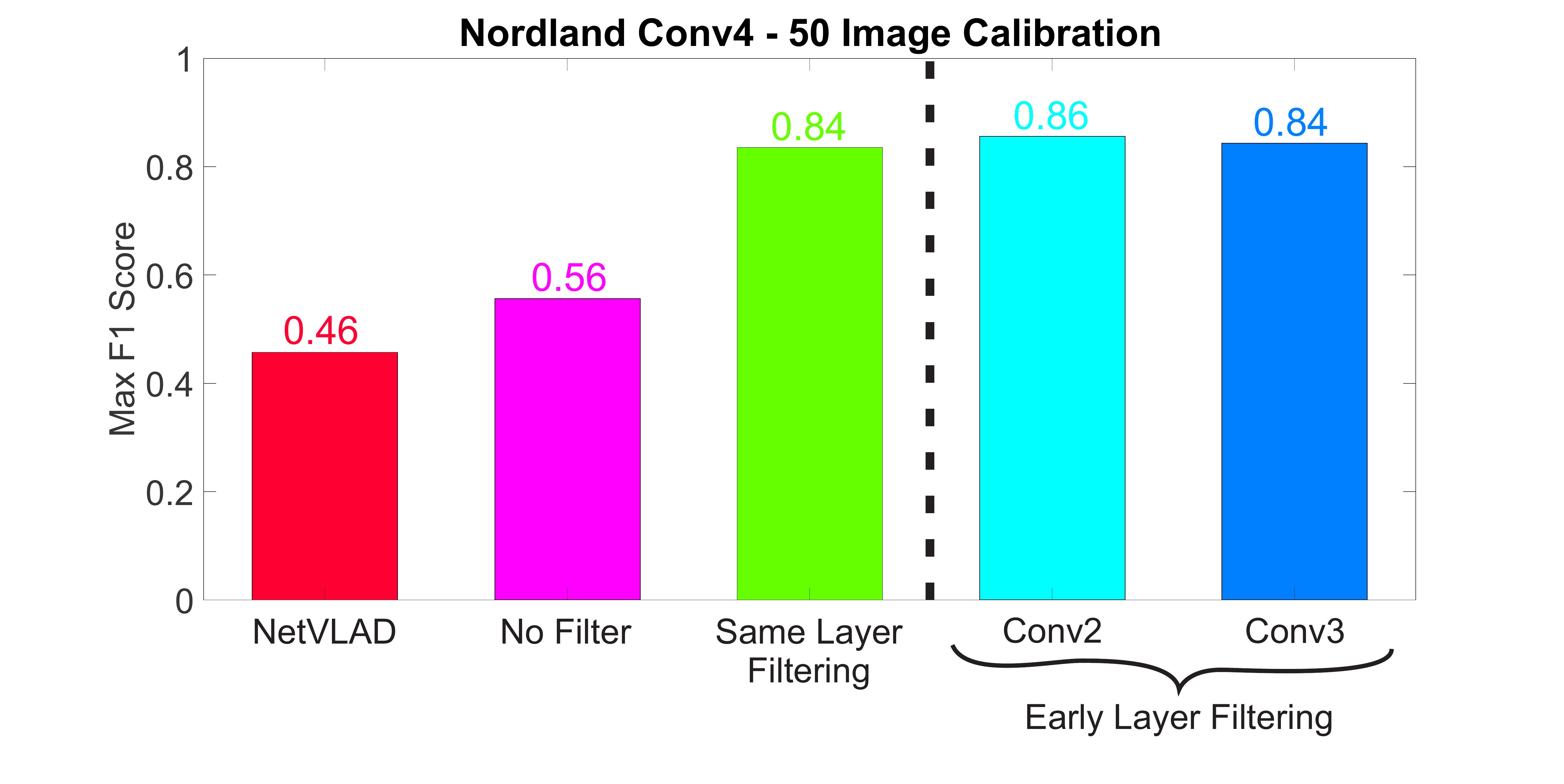}	
\caption{Maximum F1 score for Feature Map Filtering on the Nordland dataset, extracting features from Conv4. We compare the localization performance without any filtering (\emph{No Filter}), with filtering the same layer from which the image descriptor is extracted (\emph{Same Layer Filtering}), filtering either of the two previous layers (\emph{Early Layer Filtering}), and to NetVLAD.}
\label{F1BarNordConv4_50}
\end{figure}

\begin{figure}[p]
\centering
\includegraphics[width=\linewidth,trim=2cm 1.5cm 2cm 0cm,clip]{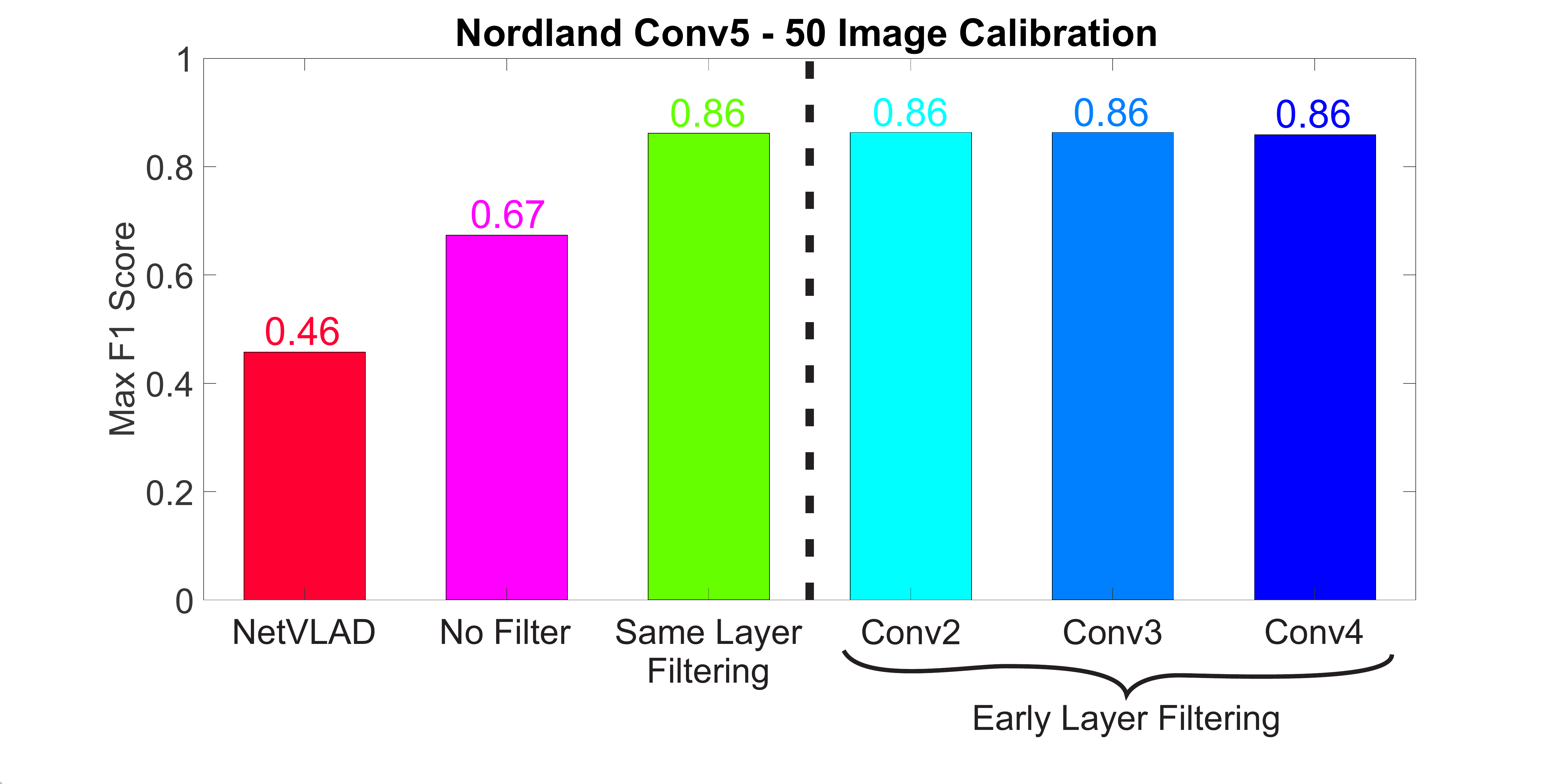}	
\caption{Maximum F1 score for Feature Map Filtering on the Nordland dataset, extracting features from Conv5. We compare the localization performance without any filtering (\emph{No Filter}), with filtering the same layer from which the image descriptor is extracted (\emph{Same Layer Filtering}), filtering either of the three previous layers. (\emph{Early Layer Filtering}), and to the comparison approach NetVLAD.}
\label{F1BarNordConv5_50}
\end{figure}

\begin{figure}[p]
\centering
\includegraphics[width=\linewidth,trim=2cm 3cm 2cm 0cm,clip]{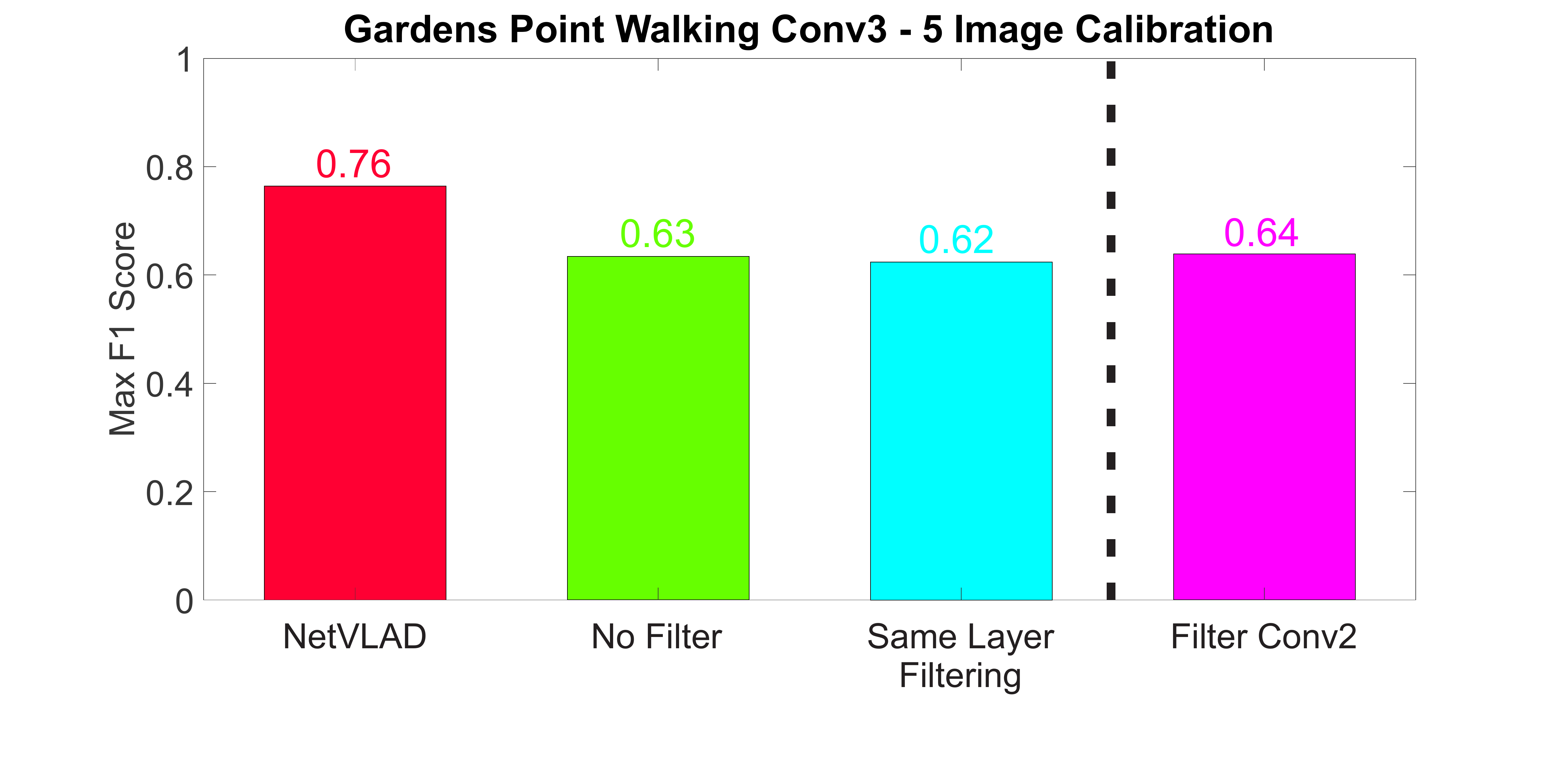}	
\caption{Maximum F1 score for Feature Map Filtering on the Gardens Point dataset, extracting features from Conv3. We compare the localization performance without any filtering (\emph{No Filter}), with filtering the same layer from which the image descriptor is extracted (\emph{Same Layer Filtering}), filtering the previous layer (\emph{Filter Conv2}), and to NetVLAD.}
\label{F1BarGPWalkConv3_5}
\end{figure}

\begin{figure}[p]
\centering
\includegraphics[width=\linewidth,trim=2cm 1.5cm 2cm 0cm,clip]{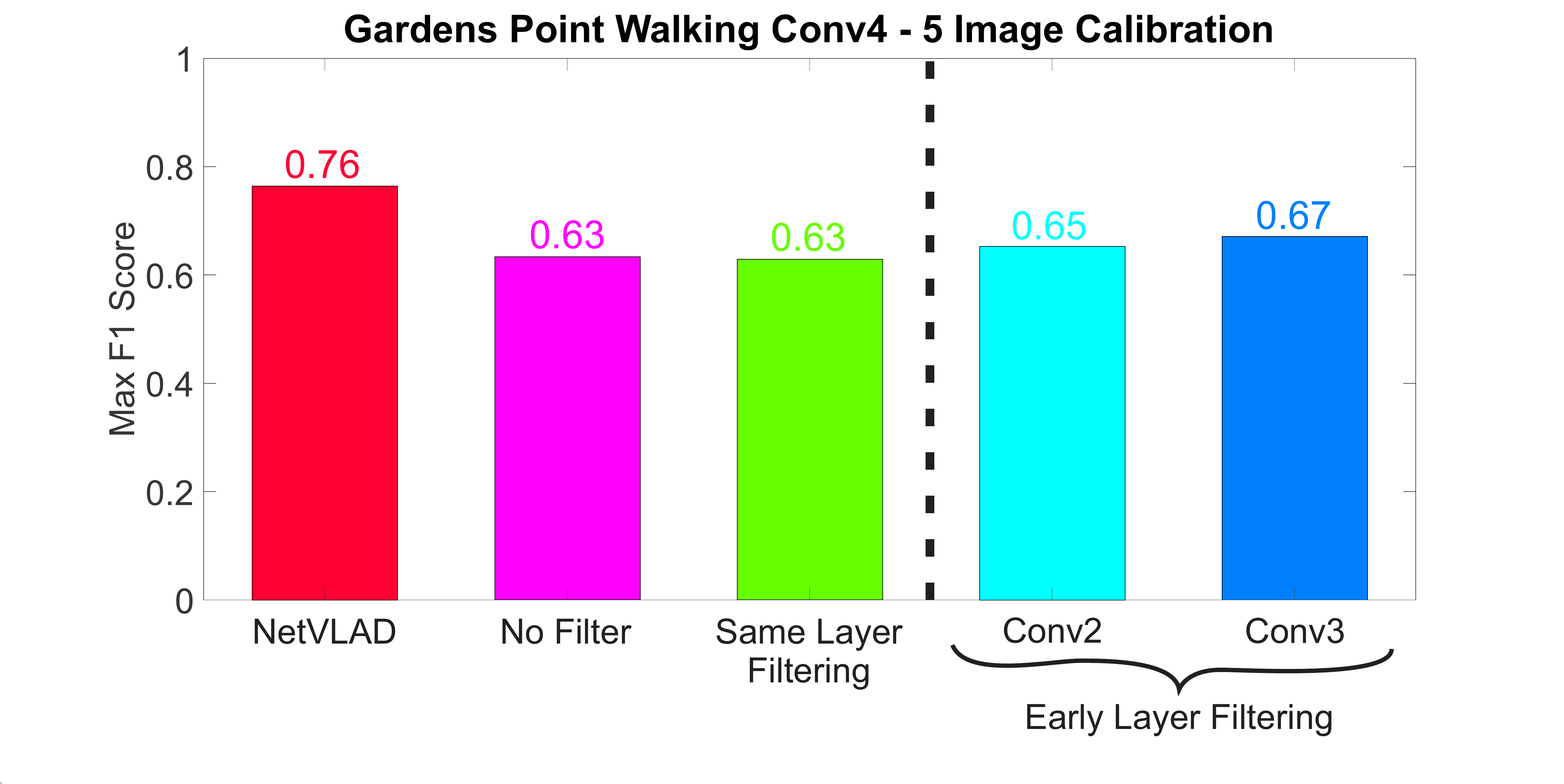}	
\caption{Maximum F1 score for Feature Map Filtering on the Gardens Point dataset, extracting features from Conv4. We compare the localization performance without any filtering (\emph{No Filter}), with filtering the same layer (\emph{Same Layer Filtering}), filtering either of the two previous layers (\emph{Early Layer Filtering}), and to NetVLAD.}
\label{F1BarGPWalkConv4_5}
\end{figure}

\clearpage


\begin{figure}[t]
\centering
\includegraphics[width=\linewidth,trim=2cm 1.5cm 2cm 0cm,clip]{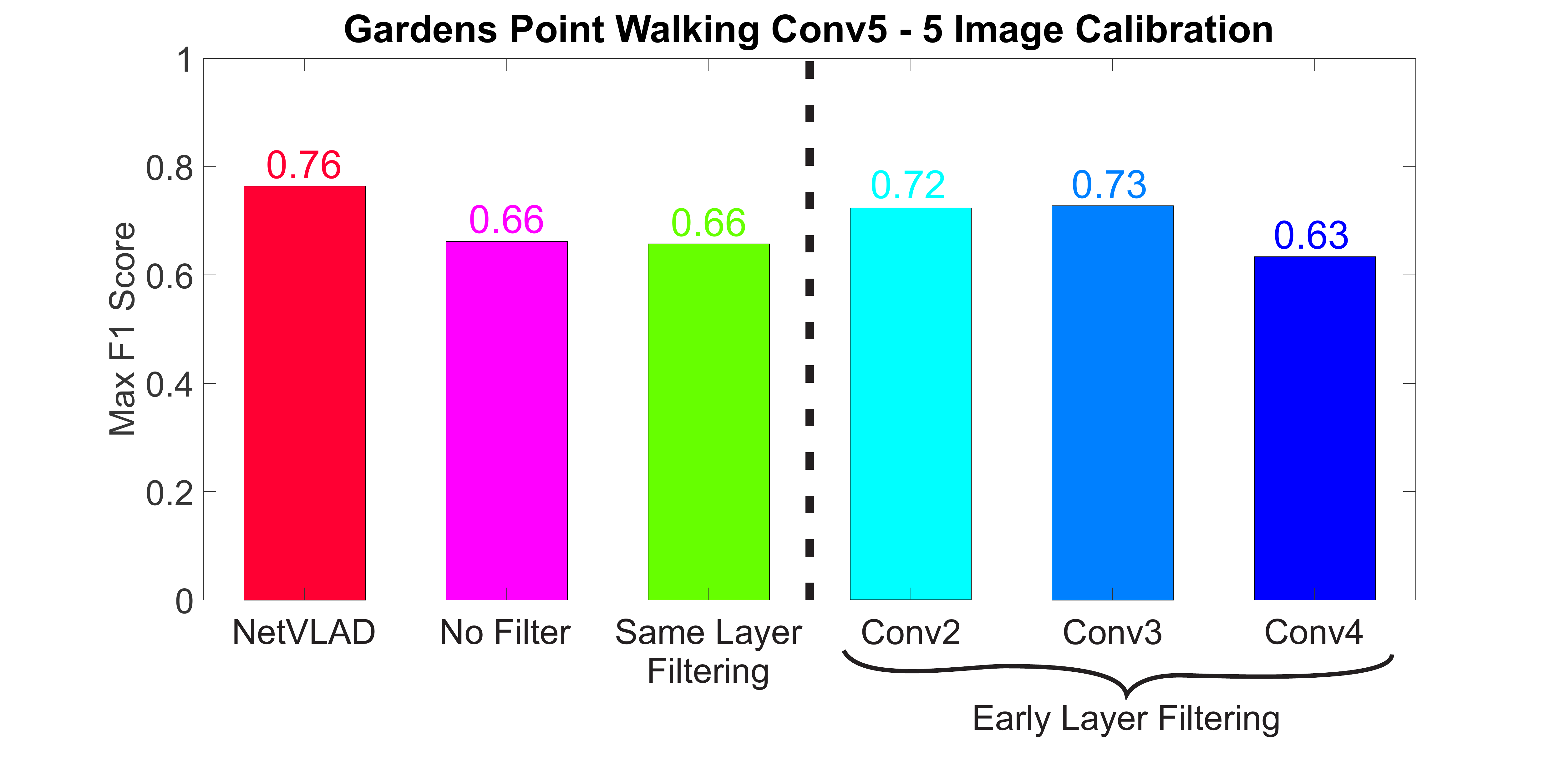}	
\caption{Maximum F1 score for Feature Map Filtering on the Gardens Point dataset, extracting features from Conv5. We compare the localization performance without any filtering (\emph{No Filter}), with filtering the same layer (\emph{Same Layer Filtering}), filtering either of the three previous layers (\emph{Early Layer Filtering}), and to NetVLAD.}
\label{F1BarGPWalkConv5_5}
\end{figure}

\subsection{Nordland}

The striking result from these experiments (Figures \ref{F1BarNordConv3_50} to \ref{F1BarNordConv5_50}) is the magnitude of improvement added with filtering, which is much greater than the other two datasets. We hypothesize that this railway dataset can be easily encoded using a set of calibration images. The environmental appearance of both summer and winter traverses changes little over the dataset, unlike the Oxford RobotCar dataset, where street lighting makes the environmental change more dynamic. 

The choice of layer to filter is mostly indeterminate on this dataset. Because there are no viewpoint variations on this dataset, the max-pooling operations between layers makes little difference to the localization performance. Therefore, once the distracting visual features are removed from any layer in the network, the choice of layer to extract a feature vector from becomes largely irrelevant.

\subsection{Gardens Point Walking}

To test our theory that early layer filtering is advantageous for viewpoint variant datasets, we used the left-hand and right-hand Gardens Point Walking traverses. As expected, the higher Conv5 layer achieved the highest localization performance, attaining a maximum F1 score of 0.73 if Conv3 is filtered first (see Figures \ref{F1BarGPWalkConv3_5} to \ref{F1BarGPWalkConv5_5}). There is a decent gap between filtering Conv3 and filtering Conv5, with a improvement in F1 score of 0.7. The improvement in F1-score using just 5 calibration images indicates that the early layer filtering process is particularly useful when moderate viewpoint variations are present. NetVLAD performs well on this dataset and beats any of our filters. NetVLAD is designed for viewpoint invariance in mind, by virtue of the learnt VLAD clustering and use of features out of the final convolutional layer. The gap between our approach and NetVLAD becomes small when Conv3 was filtered and the feature vector was produced from Conv5, using just five calibration images.

\subsection{Fully Connected Features}

We performed a final experiment, to consider using feature map filtering to optimize a feature vector formed using the first full-connected layer. We directly used the activations within the ReLu layer after the first fully-connected layer as the feature vector. We filtered Conv2 using the exact same triplet loss method on the three datasets, and show the results in Figure \ref{FCLayerF1} below. Notice how the result with a random set of filtered maps is worse than the baseline performance. This result shows that feature map filtering is beneficial because of the objective function, and not because of any inherent benefits of reducing the dimensionality of the network.

\begin{figure}[h]
\centering
\includegraphics[width=\linewidth,trim=2cm 1cm 2cm 0cm,clip]{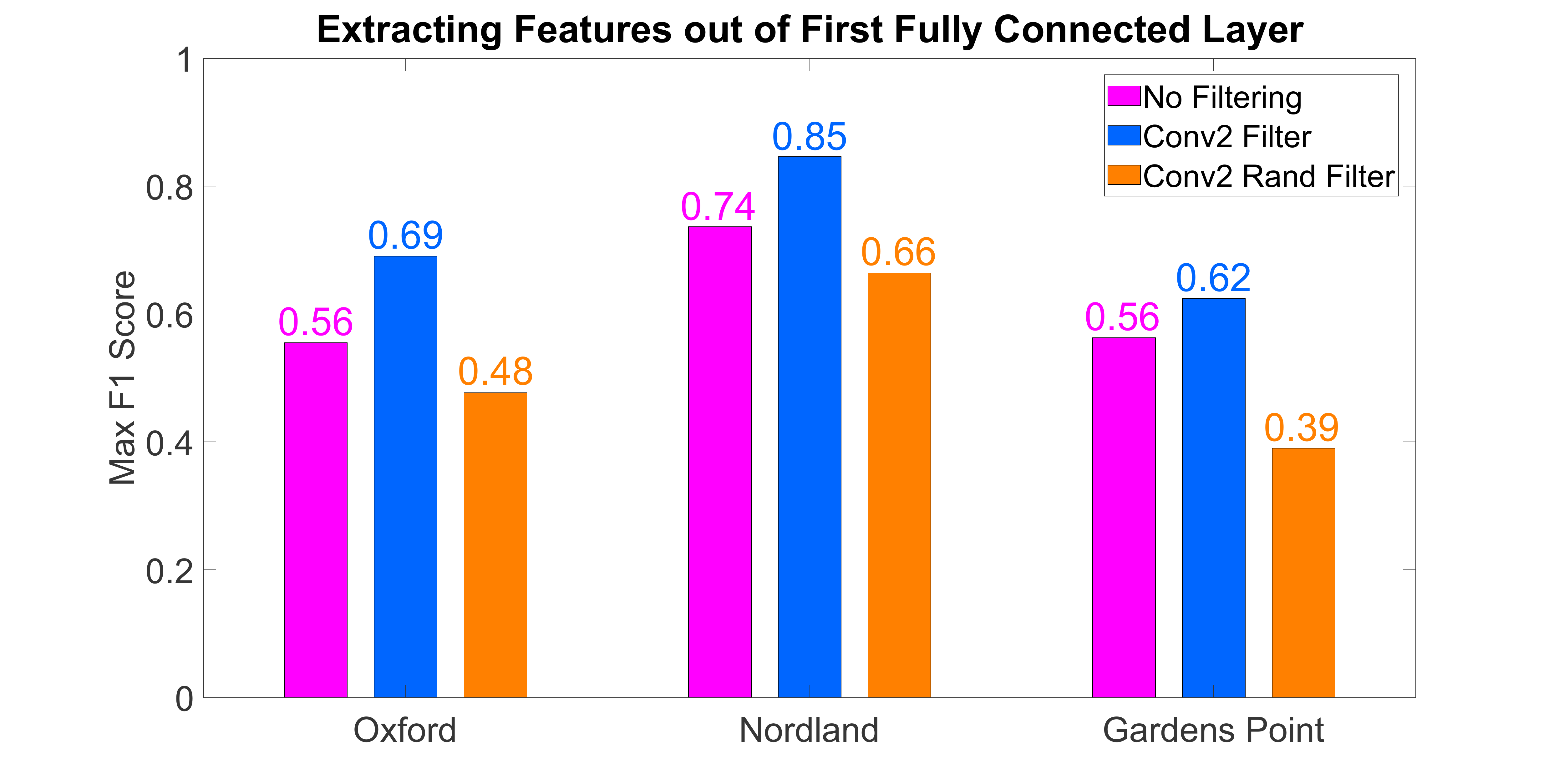}	
\caption{Maximum F1 Scores for fully connected layer features, both without any earlier filtering and also with filtering the Conv2 layer. We also include results where the same number of feature maps are removed, but with a random map selection.}
\label{FCLayerF1}
\end{figure}


\subsection{Visualization of Feature Map Filtering}

\begin{figure}[h]
\centering
\includegraphics[width=\linewidth,trim=1.7cm 6.7cm 1.7cm 1.0cm,clip]{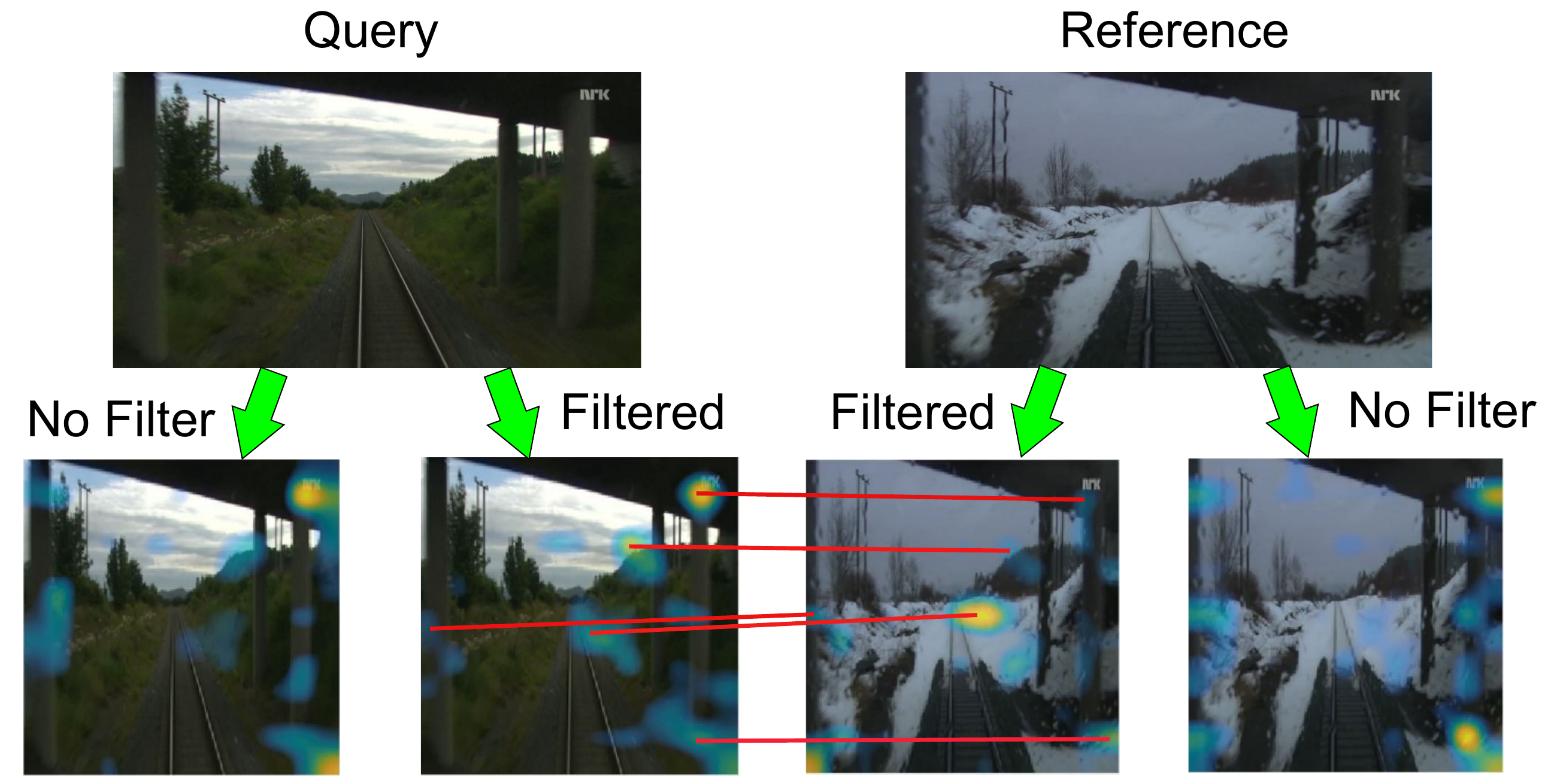}	
\caption{Visualization of Conv5 maximum activations both without filtering, and after filtering the Conv3 layer, on the Nordland dataset. The red lines indicate spatially consistent maximum activations, between the filtered maps of the reference and query images.}
\label{Visualize}
\end{figure}

By plotting the maximum activation coordinates onto a heat map, we can observe the change in activations after the network is filtered. As shown in Figure \ref{Visualize}, the filtering process effects the spatial position of the maximum activations. We found that the magnitude of these maximum activations are still different, however, the location of the maximum activations within the spatial region of the feature maps becomes more accurate, between environments. 

\section{Discussion and Conclusion}

In this paper, we presented an early filtering, late matching approach to improving visual place recognition in appearance-changing environments. We showed that CNNs tend to activate in response to features with little utility for appearance-invariant place recognition, and show that by applying a calibrated feature map filter, these distracting features are removed from the localization feature vectors. Our results indicate that filtering an \textit{earlier} layer of the network generally results in better performance than filtering the same layer that the feature vector is extracted from. We also provide a case where we filter the Conv2 layer and extract features out of the first fully connected layer, demonstrating the versatility of early layer filtering.

The experimental results show that a network layer can be severely pruned and yet continue to be run in the forward direction with coherent and effective activations in a later layer. Our approach also does not re-train after pruning, unlike many previous work in the space \cite{Li,FeatPruneFirst2017,FeatPrune2018,Lina}. Therefore this research shows that, while later layers are directly impacted by the complete removal of early features, the removal of up to 50\% of these early features does not cause a catastrophic collapse of activation strength in later layers. Note that we did find that removing more than 50\% of feature maps in an early layer dramatically increased the risk of localization instability, as the later activations experience significantly reduced activation strength. Also, removing too many feature maps during calibration risks overfitting the training data. Our results show that a small number of feature maps can be selectively pruned from an early convolutional layer, to optimize localization in the current environment. The approach is also practical from a training perspective: our results show that state-of-the-art performance can be achieved even with a single training image pair.

The work discussed here could be improved by making feature map filtering end-to-end, with the filters learnt by back-propagation. Normally a hard assignment of filtering is not differentiable, however, a soft filtering approach could be applied when training the filter. Further work will also investigate the use of feature map filtering to improve object detection and image classification. If an early layer can be filtered for the benefit of a later convolutional layer, or even a fully-connected layer, then it stands to reason that a filter could be learnt to optimize the final classifier output. 

\bibliographystyle{IEEEtran}
\bibliography{FilterEarlyMatchLate}

\end{document}